\documentclass[10pt,twocolumn,letterpaper]{article}

\usepackage[pagenumbers]{cvpr} % To force page numbers, e.g. for an arXiv version

\usepackage{graphicx}
\usepackage{amsmath}
\usepackage{amssymb}
\usepackage{booktabs}
\usepackage{multirow}
\usepackage{textgreek}
\usepackage{xcolor}
\usepackage{multicol} % Required for multi-column layouts
\usepackage{caption} 
\usepackage{tabularx}
\usepackage[accsupp]{axessibility}  % Improves PDF readability for those with disabilities.
\newcolumntype{L}[1]{>{\raggedright\arraybackslash}p{#1}}
\newcolumntype{C}[1]{>{\centering\arraybackslash}p{#1}}
\newcolumntype{R}[1]{>{\raggedleft\arraybackslash}p{#1}}
\usepackage{xr}
\usepackage[pagebackref,breaklinks,colorlinks]{hyperref}

\usepackage[capitalize]{cleveref}
\crefname{section}{Sec.}{Secs.}
\Crefname{section}{Section}{Sections}
\Crefname{table}{Table}{Tables}
\crefname{table}{Tab.}{Tabs.}

\def\cvprPaperID{5 } % *** Enter the CVPR Paper ID here
\def\confName{CVPRW}
\def\confYear{2026}

\begin{document}

%\title{EngageFSL: Student Behavior Engagement using VLM-Few-shot learning}
%\title{It’s Read from Your Behavior: Estimating student behavioral engagement from actions}
%\title{Written in His Movements: Inferring Student Behavioral Engagement from Classroom actions}
%\title{It’s Written in Their Actions: Video-Based Student Behavioral Engagement Measurement via Vision-Language and Large Language Models}
\title{Context Matters: Peer-Aware Student Behavioral Engagement Measurement via VLM Action Parsing and LLM Sequence Classification\thanks{This research was funded by NSF Award Number 2337154.}}

%\title{It's Written in His Movements: Detecting video-based student behavioral engagement from in-class behavior}% through different in-class behavior styles

\author{
Ahmed Abdelkawy\thanks{Corresponding author}\quad Ahmed Elsayed\quad Asem Ali\quad Aly Farag\quad Thomas Tretter\quad Michael McIntyre\\
University of Louisville\\
% Louisville, KY, USA \\
 {\tt\small \{ahmednady.abdelkawy,ahmed.elsayed,asem.ali,aly.farag, tom.tretter, michael.mcintyre\}@louisville.edu}
%\ead{a0nady01@louisville.edu}
% \cortext[correspondingAuthor]{Corresponding author.
}

% \author{
% Ahmed Abdelkawy $^{1}$ \\
% %\ead{a0nady01@louisville.edu}
% % \cortext[correspondingAuthor]{Corresponding author.
% % \newline \RaggedRight Email Addresses: a0nady01@louisville.edu (A. Abdelkawy), aly.farag@louisville.edu (A. Farag), islam.alkabbany@louisville.edu (I. Alkabbany), asem.ali@louisville.edu (A. Ali), tom.tretter@louisville.edu (T. Tretter)
% %a0nady01@louisville.edu
% % Institution1\\
% \and Ahmed Elsayed\\%$^{1}$
% %{\tt\small ahmed.elsayed@louisville.edu}
% \and  Asem Ali\\
% %{\tt\small asem.ali@louisville.edu}
% \and Aly Farag \\
% %{\tt\small aly.farag@louisville.edu}
% \and Thomas Tretter\\
% \and Michael McIntyre \\
% %{\tt\small [michael.mcintyre]@louisville.edu}
%  University of Louisville, Louisville, KY, USA 
%  }
% \email{
% ahmednady.abdelkawy@louisville.edu}

% }
\maketitle
%%%%%%%%% ABSTRACT
\begin{abstract}
    Understanding student behavior in the classroom is essential to improve both pedagogical quality and student engagement. Existing methods for predicting student engagement typically require substantial annotated data to model the diversity of student behaviors, yet privacy concerns often restrict researchers to their own proprietary datasets. Moreover, the classroom context, represented in peers' actions, is ignored. To address the aforementioned limitation, we propose a novel three-stage framework for video-based student engagement measurement. First, we explore the few-shot adaptation of the vision-language model for student action recognition, which is fine-tuned to distinguish among action categories with a few training samples. Second, to handle continuous and unpredictable student actions, we utilize the sliding temporal window technique to divide each student's 2-minute-long video into non-overlapping segments. Each segment is assigned an action category via the fine-tuned VLM model, generating a sequence of action predictions. Finally, we leverage the large language model to classify this entire sequence of actions, together with the classroom context, as belonging to an engaged or disengaged student. The experimental results demonstrate the effectiveness of the proposed approach in identifying student engagement. The source code will be available at 
    \url{https://github.com/ahmed-nady/context_aware_student_engagement}.
    %Current studies of student behavior analysis are either focused on recognizing student's actions or measuring engagement level by leveraging student's actions. The former require substantial annotated datasets to model the diversity of student behaviors, while the latter require substantial annotated datasets for student actions, and engagement
    %In this paper, we address the following research questions: 1) Can we build a robust video-based engagement measurement using a handful of examples? 2) Does the classroom context, represented in peers' actions, contribute to student engagement measurement? To this end
\end{abstract}
%keywords: Student behavior anlaysis, student engagement, Vision-language model, Students actions 

%%%%%%%%% BODY TEXT
\section{Introduction}
\label{sec:intro}
Students' active participation and engagement in the classroom are essential for successful learning outcomes \cite{sumer2021multimodal,farag2021toward}. Student engagement is a multidimensional construct comprising behavioral, emotional, and cognitive engagement \cite{fredricks2004school}. Behavioral engagement refers to students' self-directed actions to get access to the curriculum, such as studying and doing homework, hand and body movements while attending lectures, as well as engaging cooperatively in classroom activities \cite{andolfi2017opening}. Emotional engagement, on the other hand, refers to students' feelings about their learning experience, teachers, and peers \cite{skinner1993motivation}. This involves feelings of happiness in learning, disinterest in the subject matter, or frustration and struggle to understand. Lastly, cognitive engagement is the psychological investment in academic achievement, which is usually revealed by a student's ability to comprehend, organize, and analyze in the act of deep learning to fully understand the material \cite{chi2014icap,booth2023engagement}.   

Behavioral engagement is a key indicator of student participation and attentiveness, with a direct impact on academic performance \cite{appleton2008student}. During a lecture, students' behaviors often reveal whether they are on- or off-task. While an engaged student typically listens, takes notes, and types on a laptop, a disengaged student engages in off-task activities like playing with a mobile and frequently checking the time. %Students’ actions can indicate whether a student is on- or off-task during the lecture. Usually, an engaged student takes notes, types on a laptop, and raises hands, while a disengaged student performs off-task actions such as playing with a mobile, eating/drinking, and checking the time. 
 Conventional assessments of behavioral engagement typically depend on manual classroom observations \cite{smith2013classroom,lane2015new}. While informative, these observations are labor-intensive and subject to observer bias, which limits their scalability in large educational settings. Measuring students' engagement provides instructors with quantitative feedback, enabling them to adjust lectures and classroom activities to effectively re-engage disengaged students. An important consideration in engagement annotation is temporal resolution, which defines the time scale over which engagement is assessed. Following the standard classroom observation protocols \cite{smith2013classroom,lane2015new}, we adopt a 2-minute temporal resolution.%, i.e., the length of each video sample is set to match a 2-minute time window. 
%temporal resolution is a key factor in engagement annotation and defines the time scale at which engagement is evaluated. Therefore, the length of each video sample is set to match this annotation time window

 Automating student behavioral engagement in classrooms has received considerable attention due to advances in computer vision and deep learning. Existing studies for student behavior analysis can generally be categorized according to the target outcome into three groups: action recognition, learning analytics, and engagement measurement.  Action recognition approaches focus on identifying a particular group of students' behaviors without conducting additional analysis, providing instructors with minimal useful information. Furthermore, a considerable number of studies \cite{li2019sleep,zheng2020intelligent} formulated student behavior analysis as single- or multi-object detection, ignoring the complexity and diversity of student behaviors. Learning analytics approaches \cite{ahuja2019edusense}, which use current computer vision techniques to get a specific student's actions, gaze, and/or facial expression, provide a set of visual features but little useful information for instructors.% Specifically, a specific student's actions, gaze, and/or facial expression are usually employed in estimating the student's engagement.
 
Unlike traditional Human Action Recognition (HAR) models that operate on single-action trimmed clips, student behavior analysis demands processing continuous, untrimmed videos in which the execution and order of actions are critical. To address this challenge, research has shifted toward temporal action localization (TAL) \cite{sun2021student,yu2024raw}, which identifies and locates the start and end times of sparse, intermittent actions within long, untrimmed videos. However, TAL tends to ignore “background” segments (e.g., listening, looking to the side/back), which are important in engagement analysis. Furthermore, while Yu et al.,~\cite{yu2024raw} leveraged  a large language model (LLM) to generate a behavior analysis report, their approach does not explicitly estimate student engagement levels. To measure students' behavioral engagement, Abdelkawy et al.~\cite{abdelkawy2024measuring} represented a student's actions and their frequencies over an arbitrary time period as a histogram of actions. This histogram, along with the student’s gaze, is then used as input to an engagement classifier. However, the student’s actions within this time interval are not manually annotated, leaving the evaluation of the behavioral engagement classifier incomplete. %However, the student's actions throughout such a time interval are not manually annotated. Thus, the evaluation of the behavioral engagement classifier remains incomplete.  
%In classroom observation protocols, the interpretation of labeling a student during a time interval as engaged or disengaged relies on the student's actions, where the student is looking at, and what peers are doing as well.
% \begin{figure}
%     \centering
%     \includegraphics[width=0.95\linewidth]{images/motivation_samples.png}
%     \caption{Contribution of the context in engagement classification. The sample in the top row is annotated as disengaged even though the student was taking notes most of the time, but their peers were listening (eyes up to the instructor) all the time. The two other samples are annotated as engaged, even though the students were looking around several times for awhile. The reason is that they were looking at the student who was asking/answering question.   }
%     \label{fig:motivation}
% \end{figure}
\begin{figure}
    \centering
    \includegraphics[width=0.95\linewidth]{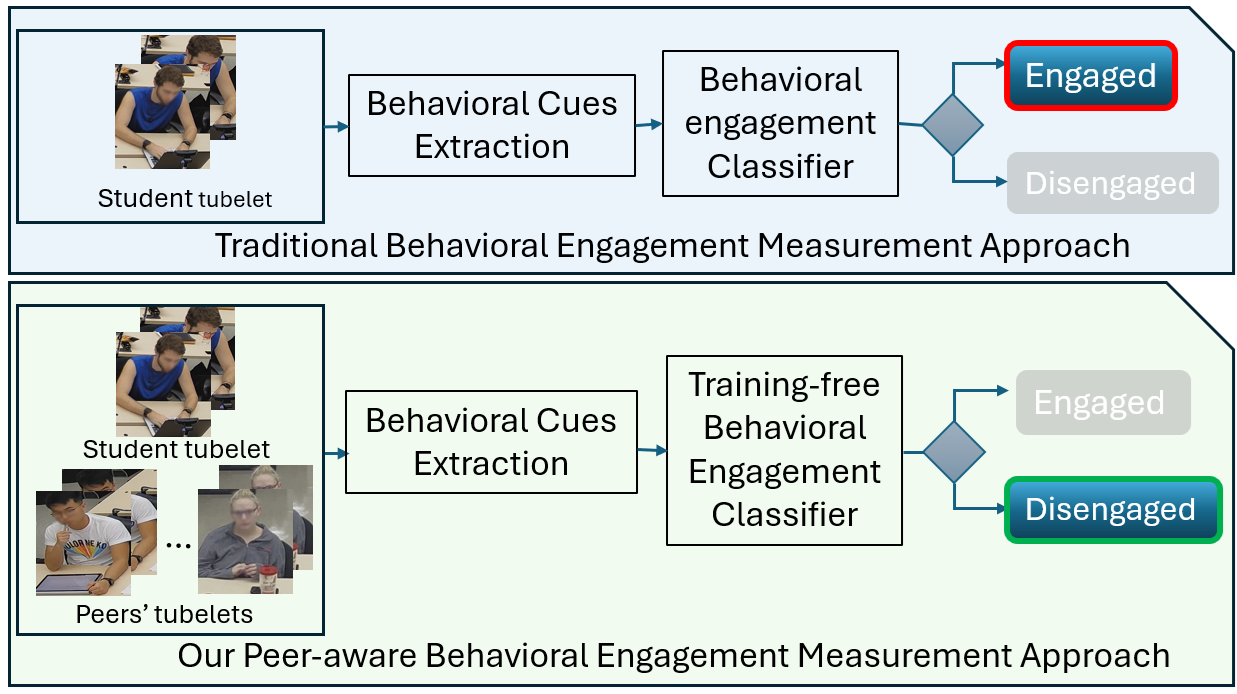}
    \caption{Context matters in engagement classification. Under traditional approaches, a student’s action sequence showing continuous typing on a laptop may be labeled as engaged. However, our peer-aware model classifies the same sequence as disengaged when surrounding students are not engaged in similar note-taking behavior.}
    \vspace{-15pt}
    %\caption{Context matters in engagement classification, since the student sequence (tube), where the student manifests typing on  the laptop, is classified as engaged through the traditional approach, while it is classified as disengaged by our approach, which takes the peers' actions in consideration.  the student typing consistently on his notebook but others are not, so it is not clear what he might be doing - likely not engaged with the class.}
    \label{fig:traditional_vs_ours}
\end{figure}
Overall, existing approaches face two key limitations: %limitations of these three types of approaches are:
\begin{itemize}
    \item Scarcity of publicly available datasets for students actions and engagement, as most studies rely on proprietary datasets due to privacy concerns. 
    \item limited consideration of classroom context, which is defined as actions performed by peers while the student of interest is performing a certain action (see \cref{fig:traditional_vs_ours}).
  
 \end{itemize}

Student engagement depends not only on their actions but also on the ongoing classroom activity and the behavior of their peers; therefore, classroom context is crucial for accurate engagement assessment. This is particularly important as contemporary pedagogy increasingly moves beyond passive lecture-based learning. The ICAP framework \cite{chi2014icap} characterizes learning activities along a Passive–Active–Constructive–Interactive hierarchy, where interactive tasks generally reflect deeper cognitive engagement. The same observable behavior can therefore signal opposite engagement states depending on context. For example, a student turning to talk to a peer during a traditional lecture probably shows disengagement, but doing the same during a peer discussion activity that is prompted shows engagement. Similarly, when a student is asked to work independently on a problem, he is engaged by focusing intently on his laptop for problem-solving. In contrast, he is disengaged if he uses his laptop for unrelated activities such as browsing social media while the instructor is lecturing.% Likewise, when a student asks a question, other students are expected to turn to listen to their peer.

To address the aforementioned limitations, we propose a novel three-stage framework for classifying behavioral engagement based on student actions. First, we perform few-shot adaptation for the vision-language model (VLM) ~\cite{rasheed2023fine} to classify human actions in trimmed videos with a limited number of labeled samples per category. Second, because the student actions are continuous, we divide each 2-minute time-resolution student video into non-overlapping segments using the sliding temporal window technique. The VLM model (from stage 1) is then used to assign a predefined label to each segment, resulting in a time-ordered sequence of predicted actions. Third, an LLM uses this sequence of actions and the classroom context as input to determine whether a student is engaged or disengaged.  
 
 %build robust engagement measurement using limited samples.
To the best of our knowledge, prior studies have not investigated whether current vision-language models (VLMs) and large language models (LLMs) can effectively recognize the student's class actions and  measure behavioral engagement based on identified actions, respectively. Accordingly, this paper addresses the following research questions:
\begin{enumerate}
    \item Compared to traditional action classification models, can VLMs accurately classify students' class actions?
    \item Which is a more accurate depiction of students' behavior: an action sequence or action histogram?
    
\end{enumerate}
 
Our contributions are two-fold:

\begin{itemize}
    \item We introduce a novel framework for measuring student behavioral engagement, which integrates VLM for action recognition and LLM for engagement classification. 
     To overcome the lack of publicly available datasets for student actions and engagement, we perform few-shot tuning for VLM to improve student action recognition, and  we leverage the LLM in zero-shot prompting for engagement classification.
    \item  Moreover, we construct a novel student behavior dataset with two components: 1) trimmed videos (13 categories, controlled environment) with annotated students' actions and 2) untrimmed videos (3 lectures, 11 students) with annotated students' engagement. The experts' annotations of this dataset include engaged/disengaged labels for each 2-minute video and dense temporal action labels for each segment within them.

\end{itemize}

\section{Related Work}
%\subsection{Action Recognition} 
%\subsection{Temporal Action Localization} 

% \begin{table*}[htbp]
% \centering
% \caption{Summary of recent works on automatic student engagement analysis.}
% \begin{tabular}{p{1.7cm}  p{2cm} p{2cm} p{2.5cm} p{3cm} p{3cm}}
% \hline
% \textbf{Method} & \textbf{Temporal Unit} & \textbf{Modalities} & \textbf{Output} & \textbf{Model} \\
% \hline
% \cite{Sheng2025}  & One frame & Classroom RGB image & Per-frame student actions & YOLOv8-based CNN with MLKCM and PFOM modules \\

% \cite{Yu2024} & Video sequences & Classroom RGB + 2D skeletons & LLM-generated behavior reports & 2D HPE + Temporal Action Detection + LLM  \\
% \cite{abdelkawy2024measuring}& 2-minute-long video & 2D skeletons + head gaze &engagement level & Random forest, SVM \\
% \cite{Marquez-Carpintero2025} & Frame / short seq. & RGB image, IMU, HRV & Attention level (1–5), engagement label & VLM, MLLM \\
% \hline
% \end{tabular}
% \label{tab:engagement_lit}
% \end{table*}

\subsection{Student Engagement Measurement}
The emerging and promising real-time engagement measurement approaches, enabled by advances in computer vision and machine learning, allows tracking a student's engagement at fine-grained temporal resolutions on the order of seconds or minutes through discrete, objective indicators. Existing research of student behavior analysis in classrooms can be grouped into three categories: student action recognition, learning analytics, and engagement measurement. While student action recognition approaches focus on the student's actions from trimmed videos or a single image, learning analytics approaches rely on facial expressions, student actions, and/or head gaze. For example, Sheng et al.~\cite{Sheng2025} proposed an enhanced YOLOv8-based detector for localizing and classifying student behaviors from single classroom images. Their framework incorporates a Multi-Scale Large Kernel Convolution Module (MLKCM) to extract multiscale spatial features and a Progressive Feature Optimization Module (PFOM) that segments the channel dimension of the input feature map for improved feature refinement. Ahuja et al. \cite{ahuja2019edusense} presented a classroom sensing system (EduSense) that provides a set of features, including detection of hand raise, hand on face, cross arms, smile, head orientation, and speech duration for both the teacher and the students. These features are extracted from students'  skeleton joints and facial landmarks due to the presence of much visual noise, occlusion in the classroom, and privacy preservation. %The authors employed students' skeleton joints and students' facial landmarks to extract these visual features due to the presence of much visual noise, occlusion in the classroom, and privacy preservation. The student's actions are recognized frame-wise, which leads to a high false-positive rate in recognition, especially for hand raising.
 Line et al.\cite{lin2021student} recognized student behavior in the classroom based on upper body skeleton joints. They first extract students' poses using bottom-up pose estimation (OpenPose) and remove the incorrect connection of students' joints using YOLOv4 object detection, then generate a feature vector using normalized joint locations, joint distances, and bone angles to represent student posture and use a four-fully-connected-layer neural network to classify the student feature vector into four classes. The limitation of works \cite{ahuja2019edusense,lin2021student} is that student actions are recognized from single images rather than video segments, thereby ignoring temporal dynamics of the actions. Moreover, Line et al.\cite{lin2021student} consider actions with different engagement levels, such as writing and playing with a phone, as one class (bowing). 
 %The main limitations of their framework are that it recognizes student actions from single images rather than video segments, and it considers actions with different engagement levels, such as writing and playing with a phone as one class (bowing). 

Student engagement studies, which employ student actions and/or head gaze to measure the student behavioral engagement, provide teachers with actionable insights. In \cite{Yu2024}, the authors proposed a framework that processes classroom videos by first estimating students' 2D poses per frame and then applying a Temporal Action Detection (TAD) \cite{Filtjens2024} to produce a temporal sequence of actions for each student. These per-student action sequences are subsequently fed into an LLM to generate behavioral reports. The TAD module was trained on the NTU RGB+D dataset \cite{Shahroudy2016}, using action classes relevant to the classroom environment. However, their approach fails to provide explicit engagement labels and does not account for the peer-aware context when generating a textual behavioral report that includes each student's actions analysis. %this approach has two main limitations. First, it assumes that each student’s behavior is independent of others, ignoring the peer-aware context. Second, the model does not predict explicit engagement labels; instead, it generates textual behavioral reports that require human interpretation to extract meaningful insights. 
Abdelkawy el al.\cite{abdelkawy2024measuring} utilized a random forest classifier to detect whether the student is on-task or off-task based on a histogram of actions, together with head gaze. Specifically, each student's 2-minute video was divided into overlapping segments using the sliding temporal window technique; the student skeleton sequence of each segment is assigned a predefined action label using 3D CNN model and then the histogram is created from these identified actions. Nevertheless, the student's actions during the given period are not manually annotated. Consequently, the evaluation of the behavioral engagement classifier is still incomplete. Unlike these studies, we analyze the entire 2-minute video to get the student's sequence of actions, which are used for engagement measurement. Moreover, we constructed a dataset that has dual expert annotations: engaged/disengaged for each 2-minute video and its corresponding frame-wise action labels to measure the performance of the engagement classifier. We also investigated the few-shot adaptation of the vision-language model for student action recognition, which learns to differentiate between action classes with a few training samples to overcome the challenge of the scarcity of annotated data required to model the diversity of student behaviors. 

%The three main visual indicators commonly used in this approach are eye gaze \cite{Carter2020}, facial expressions \cite{Alkabbany2023}, body actions \cite{abdelkawy2024measuring}. Facial expression and eye gaze analysis typically require complex classroom setups, for instance, one camera per student. In contrast, body actions can be captured reliably for the entire classroom using only one or two cameras, even in large classrooms, which makes them a practical and scalable indicator. Accordingly, body actions are the sole indicator we employ to assess students’ behavioral engagement in the classroom. 

%However, the study focused solely on behavior detection and did not establish connections between the identified behaviors and validated engagement measures, nor did it examine whether behavior frequencies or temporal dynamics could predict engagement levels. 

% due to not incorporating action dynamics.

\begin{figure*}[htb]
  \centering
   \includegraphics[width=0.95\linewidth]{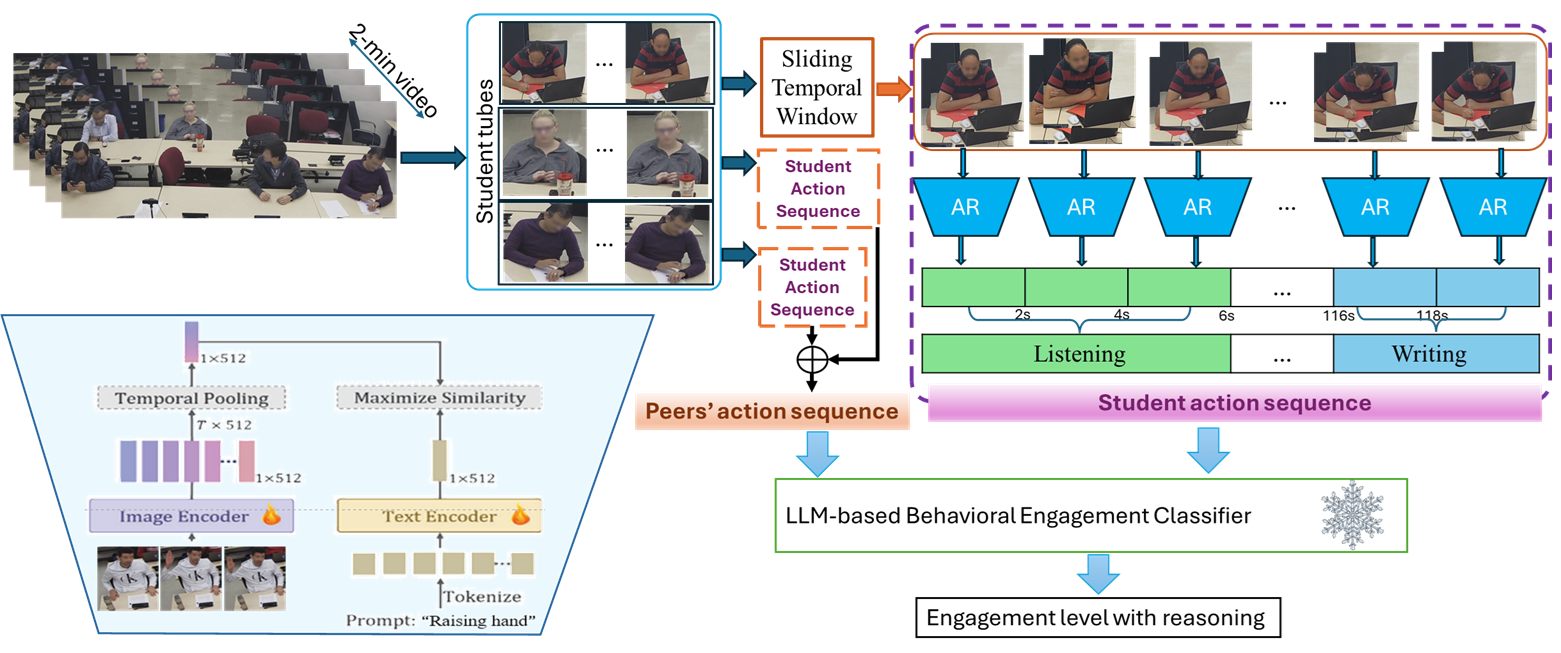}
   \caption{Overview of the proposed framework. Students' tubelets are extracted from a two-minute video, and each student's tube is divided into non-overlapping segments using a sliding temporal window technique. The action sequence is then produced by the fine-tuned VLM model. LLM uses the generated action sequence to classify engagement after integrating it with peers' action sequences.}
   \vspace{-15pt}
   \label{fig:framework}
\end{figure*}

\vspace{-5pt}
\subsection{Video Action Recognition}
\vspace{-5pt}
Action recognition aims to assign a predefined action label to a trimmed video, which contains a single action. The majority of existing action recognition approaches are based on RGB video streams. Existing RGB video-based methods can be categorized into four classes based on the network architecture: two-stream 2D convolutional neural network (CNN), 3D CNN, compute-efficient, and transformer-based methods \cite{abdelkawy2025epam}. Two-stream 2D CNN-based methods \cite{simonyan2014two,wang2016temporal} learn the appearance and motion features separately using two 2D CNN streams. 3D CNN-based methods \cite{carreira2017quo,feichtenhofer2019slowfast} are introduced to learn spatiotemporal features by temporally extending the convolution and pooling kernels of a 2D CNN. However, 3D CNNs are computationally intensive. To achieve the computation/accuracy tradeoff, compute-efficient networks \cite{tran2018closer,feichtenhofer2020x3d} are introduced. Transformer-based methods \cite{bertasius2021space,arnab2021vivit} adapt the de facto vision transformer to model the long-term temporal dynamic in a video sequence. While most of these methods are unimodal, ignoring the semantic information of action categories, works such as ActionCLIP \cite{wang2021actionclip} and XCLIP \cite{ni2022expanding} adapt the CLIP to vidoe action recognition through fusing the vision-language representation of CLIP with introduced temporal modeling components. Rasheed et al.\cite{rasheed2023fine} found that these introduced components can negatively impact CLIP’s generalization capability, whereas full fine-tuning of CLIP on the Kinetics-400 dataset yields competitive performance. In this work, we adapt the work of \cite{rasheed2023fine} in a few-shot setting to boost the performance of student classroom action recognition.
\vspace{-5mm}
\subsection{Large Language Models}
\vspace{-1mm}
Large language models (LLMs) are Transformer-based models with hundreds of billions of parameters, trained on large-scale text corpora (e.g., GPT-3 \cite{brown2020language}, LLaMA \cite{dubey2024llama}, Gemma \cite{team2024gemma}, etc.). They exhibit generalization abilities, enabling them to perform tasks in zero-shot settings without requiring task-specific training examples.
While several works have employed LLMs in education for multiple-choice question generation and assignment grading, Yu et al. \cite{yu2024raw} utilized an LLM to generate a pedagogical report based on the structured temporal action information for each student, including action categories and their timestamps. In contrast, our work uses an LLM to assign an engagement level to each student's sequence of actions. 

\section{Method}
%\subsection{Problem formulation}

\subsection{Overview of the proposed method}
The proposed framework consists of three main steps:  segment-wise action recognition, VLM action parsing, and engagement classification (\cref{fig:framework}). First, a vision-language model is adapted in a few-shot manner using a small set of labeled trimmed student action clips to improve recognition on continuous student videos. Second, the student's untrimmed video is split into non-overlapping segments using a sliding-window technique, and the fine-tuned VLM assigns an action label to each segment. Consecutive identical predictions are merged to form an ordered action sequence that preserves both action identity and duration. Finally, this action sequence, along with the classroom context, is provided to an LLM to determine whether the student is engaged or disengaged. 

\begin{figure}
    \centering
    \includegraphics[width=0.9\linewidth]{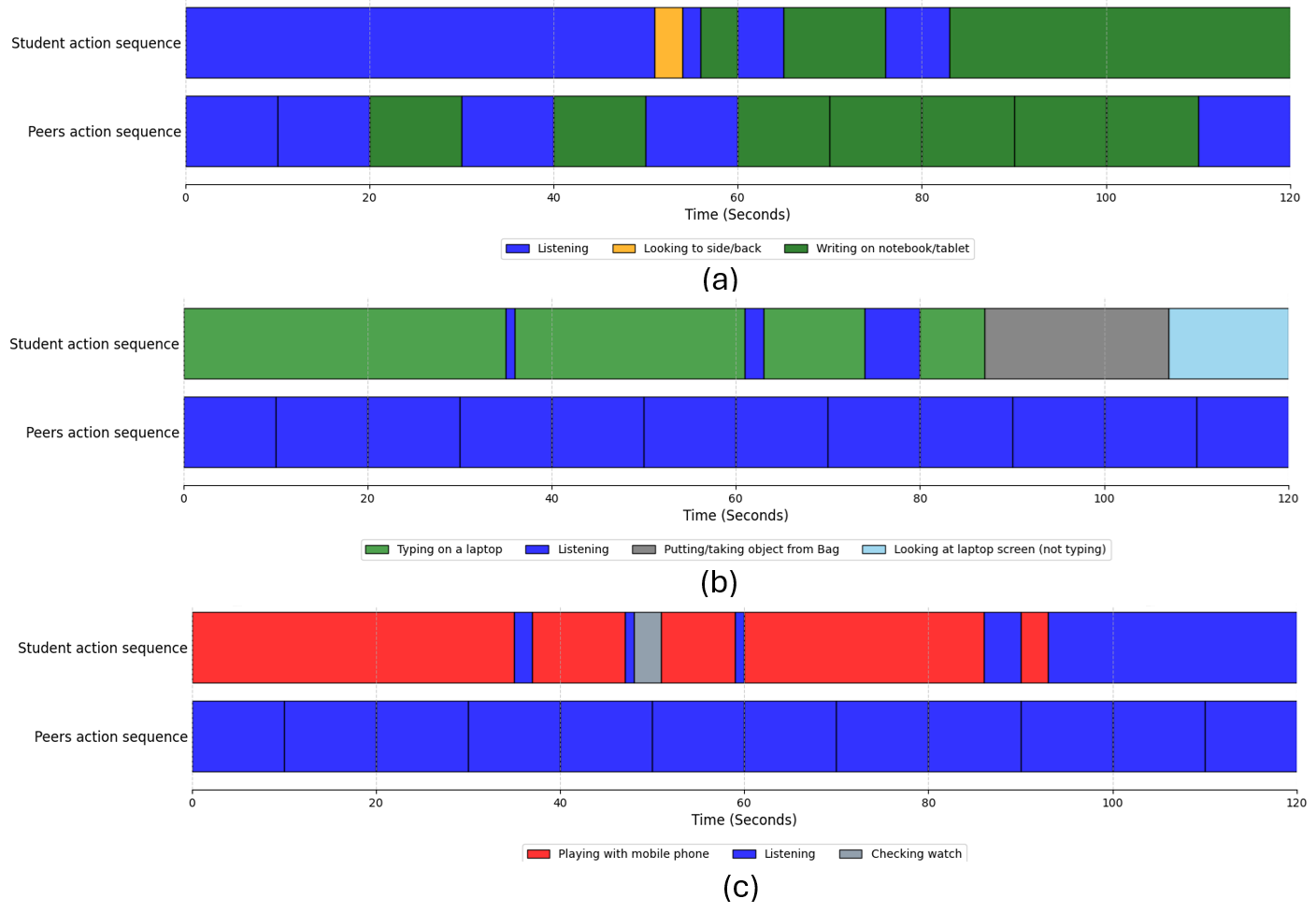}
    \caption{a) for engaged student, b, c) for disengaged student}
    \vspace{-10pt}
    \label{fig:placeholder}
\end{figure}

Representing student's behavior during a 2-minute interval as a temporal sequence rather than a histogram, where the duration of each action is accumulated, allows the model to capture not only how long each action occurs but also their ordering, which is shown to be crucial for interpreting engagement. For example, a student may still be deemed engaged if they take notes for the majority of the time period but play on their phone for one continuous 25-second block (e.g., at the beginning or end). However, a student is considered disengaged if they alternate between taking notes and making frequent quick phone checks because this pattern of frequent disruption shows an inability to maintain cognitive focus on the task. Therefore, the temporal sequence and context of actions more accurately reflect engagement than the total amount of time spent in each action (e.g., via a histogram).
Furthermore, when representing student action as a sequence, together with peers' action sequences, the LLM predictions are easier to interpret. Fig.\ref{fig:placeholder} shows examples of students' actions aligned with classroom actions for engaged (a) and disengaged (b, c) students.
\vspace{-5pt}
\subsection{ Actions dictionary}
\begin{figure}
    \centering
    \includegraphics[width=0.95\linewidth]{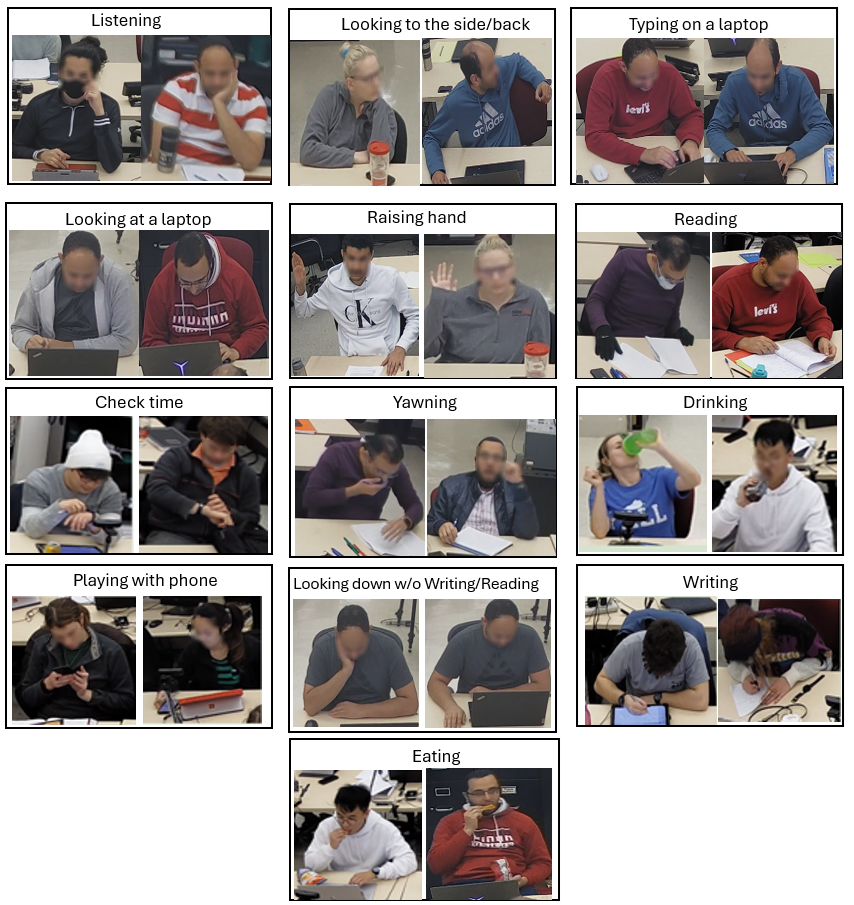}
    \caption{Samples of students' actions during class time.}
    \vspace{-15pt}
    \label{fig:actions}
\end{figure}

%Since this study is conducted on college students in introductory engineering STEM classes, the actions they perform during classroom time, from which student behavior engagement can be estimated, are limited. 
Because this study focuses on college students in introductory engineering STEM courses, the range of classroom actions from which behavioral engagement can be inferred is relatively limited. We adopted the classroom observation protocols in \cite{lane2015new,abdelkawy2024measuring}, and identified the following 13 actions:  Eating meal/snack, Writing on notebook/tablet, Typing on a laptop, Playing with mobile phone, Looking to the side/back, Looking down w/o reading/writing, Looking at laptop screen (not typing), Raising hand, Checking time, Reading, Drinking, Yawning , and Listening.
%  \begin{itemize}
%     \begin{minipage}{0.69\linewidth}
%   \item Eating meal/snack, 
%     \item Writing on notebook/tablet,
%         \item Typing on a laptop,
%         \item Playing with mobile phone,  
%         \item  Looking to the side/back, 
%          \item Looking down w/o reading/writing,
%         \item Looking at laptop screen (not typing).      
%   \end{minipage}
%   \begin{minipage}{0.28\linewidth}        
%          \item Raising hand, 
%         \item Checking time, 
%         \item Reading,        
%         \item Drinking,
%         \item Yawning, 
%         \item  Listening,      
%         \item 
%   \end{minipage}
% \end{itemize}
   Fig.~\ref{fig:actions} shows samples of such actions. 

\subsection{Student action recognition}
%on the Kinetics-400 dataset
We investigated few-shot learning for action recognition using vision-language models (VLMs). In particular, we use Video Fine-tuned CLIP (ViFi-CLIP) \cite{rasheed2023fine} because of its efficiency and simplicity. %The ViFi-CLIP adapted CLIP to the video domain through fine-tuning both the CLIP image and text encoders. 
For a video segment $S_{i} \in \mathbb{R}^{TxHxWxC}$, the embeddings of its T frames are obtained using the CLIP image encoder and then average-pooled to get the video-level representation $v_i \in \mathbb{R}^{D}$ . The corresponding class label $C_i$ is embedded in a prompt (e.g., ‘a photo of a $<$category$>$’) and passed through the CLIP text encoder to produce a text embedding $t \in \mathbb{R}^{D}$. We fine-tuned the ViFi-CLIP model on the student actions dataset in a few-shot setting with 16 video clips per action category by maximizing the cosine similarity between a batch of video-level embeddings $v_i$ and their corresponding text embeddings $t_i$ using cross-entropy loss. Additionally, we incorporate entropy regularization into the loss function to enhance generalization and mitigate overfitting by discouraging overly confident, incorrect predictions.
\begin{equation}L_{total} = - \frac{1}{N} \sum_{i}^{N} log(p_{i}) + \frac{1}{N} \sum_{i}^{N}(-\sum_{k=1}^{C} p_{i,k}  log (p_{i,k})) \end{equation} where $ p_{i}= \frac{exp(sim(v_{i},t_{i})/\tau}{\sum_{j} exp(sim(v_{j},t_{j}/\tau)}$ 
%\begin{equation}
%    p_{i}= \frac{exp(sim(v_{i},t_{i})/\tau}{\sum_{j} exp(sim(v_{j},t_{j}/\tau)}
%\end{equation}

%ViFi-CLIP bridges the modality gap between images and videos by optimizing both the CLIP image and text encoders using the Kinetics-400 dataset.  The CLIP text encoder is used to encode the class labels, and the CLIP image encoder is used to obtain the frame-level embeddings for a video sample, which are then average-pooled to obtain the video-level representation. The class label with the highest cosine similarity is then selected as a label for the specified video after the cosine similarity between the video-level representation and class label embeddings is calculated.}
\vspace{-3pt}
\subsection{Student engagement classification}
\vspace{-3pt}
To determine whether a student is engaged or disengaged, we leverage the LLM as a classifier in zero-shot prompting. Following the prompt-based learning paradigm, the LLM is conditioned on a prompt $x_{prompt}$ to generate a predefined textual label $y \in Y$ (e.g., engaged, disengaged). The prompt $x_{prompt}$ is constructed by concatenating a \textbf{task description} $x_{desc}$ and \textbf{input} $x_{input}$ as \{$x_{desc}$;$x_{input}$\}. An example of the task description $x_{desc}$, which describes the engagement classification based on the student's actions and classroom context, is provided in supplementary material \ref{sec:used_prompts}. % shown in \cref{supp-sec:used_prompts}.% \ref{sec:context_based_prompt}. %Fig.~\ref{fig:context_based_prompt}
An example of the input $x_{input}$, which includes the student action sequence with the duration of each action, along with the classroom context, represented in the majority of peers' actions during the time interval, can be represented as:
\vspace{-5pt}
\begin{itemize}
  \item \texttt{Student actions: \{ writing on tablet (00:00-00:20); listening (00:20-01:05; etc.\}} \vspace{-5pt}
        \item \texttt{Classroom context: \{listening (00:00-00:10); writing (00:10-00:50); etc.\}}    
\end{itemize}

\vspace{-10pt}
\section{ Experiments}
\vspace{-5pt}
\subsection{Experimental Setup}
\textbf{Benchmarks.} 
The lack of a public benchmark for classroom engagement analysis makes it necessary to create one. So, we created a behavioral engagement dataset with two subsets of annotated videos—one for student actions and the other for student engagement—in order to independently assess each step of the proposed framework.  All videos were recorded from a wall-mounted 4K camera at 15 fps in classrooms of 6–11 students. For the student action dataset, we recorded 7 subjects performing 13 classroom-related actions in a controlled environment. This resulted in 208 trimmed clips for training (16 per action) and 46 trimmed video clips for testing across the 13 action categories. This trimmed video dataset is designed to evaluate the proposed action recognition stage (Stage 1). The student engagement dataset comprises videos from three half-hour lectures involving eleven students. Each video is segmented into 2-minute clips per student, yielding a total of 455 video clips. Experts in psychology and education labeled each two-minute clip as either engaged or disengaged, resulting in 400 engaged and 55 disengaged samples. Each two-minute video segment is also temporally partitioned manually, and every resulting segment is assigned a predefined action label. The entire student engagement dataset is used as a test set to evaluate the remaining stages—proposed temporal action segmentation (Stage 2) and engagement classification (Stage 3)—since both stages are training-free: Stage 2 employs a temporal sliding window, while Stage 3 uses an LLM with zero-shot prompting. %The entire student engagement dataset is utilized as a test set to evaluate the remaining stages of the pipeline: proposed temporal segmentation (Stage 2) and engagement classification (Stage 3) since Stages 2 and 3 are training-free, i.e., Stage 2 employs a temporal sliding window and Stage 3 uses an LLM in zero-shot prompting.

\textbf{Implementation details.}
We implemented the proposed framework in PyTorch on a workstation with two NVIDIA TITAN RTX GPUs (24 GB each). The VIFI-CLIP model (ViT-B/16 backbone) is fine-tuned in a few-shot setting with \textbf{K}=16 samples per class, using a batch size of 32 and a learning rate $2\times 10^{-6}$ for 50 epochs.
We use the AdamW optimizer with a weight decay of 0.001 and apply cosine annealing for learning rate scheduling. Each video clip is represented by 16 uniformly sampled frames, which are resized to $224 \times 224$. For VLM-based action parsing, both the sliding window size and stride are set to 3 seconds (45 frames), resulting in 40 non-overlapping segments for each 2-minute video. For an LLM-based engagement classifier, we set the temperature to 0.1 to ensure a deterministic and non-creative output.  %the sliding window size and stride are both set to 3 seconds (45 frames), producing 40 sliding-window segments per 2-min video. 

Our proposed framework consists of distinct tasks: action recognition, action parsing, and engagement classification. Each stage is evaluated with task-appropriate metrics. For action recognition, we report the top-1 accuracy. For action parsing, we adopt standard temporal action segmentation metrics, which fall into two categories: frame-based and segment-based metrics \cite{ding2023temporal}. The frame-based metric, Mean over Frames (MoF), measures the ratio of correctly predicted frames to the total number of frames. However, the MoF metric suffers from two major problems. First, it is unreliable for imbalanced datasets, where long-duration action classes can dominate the sequence. Second, it fails to capture segment quality. A high MoF can still be achieved even if the action segments are fragmented (over-segmented). To account for the over-segmnetation phenomenon, we also report two segment-based metrics: Edit Score and $F1@\tau$. The Edit Score evaluates the similarity between the predicted and ground-truth action sequences without requiring exact frame alignment, capturing the correctness of action order. The $F1@\tau$ score measures the Intersection-over-Union (IoU) between predicted and ground-truth segments under threshold τ, counting a segment as a true positive only when its IoU exceeds $\tau$ (with $\tau \in \{10, 25, 50\}$). Multiple predictions overlapping a single ground-truth segment result in one true positive, and the rest are counted as false positives. %These segment-based metrics more effectively assess over-segmentation and temporal coherence in the predicted sequences.
Finally, we evaluate the engagement classification stage using precision, recall, and F1 score. These metrics are selected over simple accuracy to effectively assess performance in the presence of class imbalance between the engaged and disengaged labels.

\subsection{ Evaluation of the proposed framework stages}
\textbf{Stage \#1: Action Recognition}.
 We evaluated state-of-the-art video-language models, including XCLIP \cite{ni2022expanding}, TC-CLIP \cite{kim2024leveraging}, and ViFi-CLIP \cite{rasheed2023fine} on our student classroom action test set (46 instances) under zero-shot and few-shot protocols.
For the zero-shot setting, the ViFi-CLIP achieved 52.1\% top-1 accuracy, which is far from satisfactory. We attribute this poor performance to the model's fine-tuning on general web videos (Kinetics-400). While this bridges the image-video gap for general tasks, it fails to capture the subtle, fine-grained differences in our domain, such as differences between reading and writing. For the few-shot setting, as shown in \cref{table:comparison_ar_models}, we note that the performance improves for all methods as the number of shots increases.  Across most values of K, ViFi-CLIP achieves the highest accuracy, with particularly noticeable gains in the low- and mid-shot regimes, demonstrating that it can be effectively adapted to our specialized domain with minimal data. For example, ViFi-CLIP achieves gains of 4.5\%,2.2\%, and 4.6\%, compared to XCLIP at k=2, K=4, and K=8, respectively.
\begin{table}
\caption{Comparison of action recognition models.}
\vspace{-15pt}
\label{table:comparison_ar_models}
\begin{center}
{\footnotesize
\begin{tabular}{|ccccc|} 
 \hline
 \multirow{2}{*}{Model} & \multicolumn{4}{c|} {Student Action Dataset}\\  \cmidrule(l){2-5}
      &K=2& K=4&K=8&K=16 \\  \hline
%Vanilla CLIP \cite{radford2021learning} & \textbf{52.1\%}& 52.1\% &52.1\%  &52.1\%  \\  \hline
TC-CLIP \cite{kim2024leveraging} &52.2\%  & 52.2\%  & 80.4\%  & 93.5\%  \\ \hline
XCLIP \cite{ni2022expanding}&  43.2\%&79.6\%  & 88.6\% & 97.9\%  \\ \hline
ViFi-CLIP \cite{rasheed2023fine}& 47.7\% & \textbf{81.8.0}\% &\textbf{ 93.2\%} &\textbf{  97.9\% }\\ \hline

\end{tabular}
}
\end{center}
\end{table}

 \textbf{Stage \#2: Temporal action segmentation (TAS)}. \cref{table:comparison_segmentation} shows the comparison of Gemini 2.5 Pro and Gemini 2.5 Flash, which are  multimodal large language models (MLLMs), with the proposed VLM action parsing on the student engagement dataset, which provides dense action labels for each 2-minute video. For MLLMs, we curated a prompt describing the task to be performed on the given 2-min video (see supplementary material \ref{sec:TAC_prompt} for details). %\ref{sec:TAC_prompt}.%\cref{fig:action_segm_prompt}.

Dense student action labeling is challenging due to fine-grained, visually similar actions (e.g., `looking at laptop' vs. `looking down') and cluttered environments. The task is further complicated by inter-student occlusion, where objects from one student's space appear in another's, creating ambiguous visual data. It is noticeable from \cref{table:comparison_segmentation} that Gemini-2.5-pro achieves higher MoF, edit score, and F1 score compared to both Gemini-2.5-flash and the proposed VLM action parsing, while the proposed VLM action parsing outperforms Gemini-2.5-flash across all three metrics. Moreover, the proposed VLM action parsing is more consistent in identifying the on- and off-task actions than both Gemini MLLMs. This leads to accurate engagement classification, as shown next. 

 \begin{table}[t]
\caption{Comparison of temporal action segmentation using VLM-based, Gemini-2.5-Flash, and Gemini-2.5-Pro-based.}%Writing notes/reading means the two are merged into one class
\vspace{-15pt}
\label{table:comparison_segmentation}
\begin{center}
{\footnotesize
\begin{tabular}{|C{2cm}|C{0.9cm}|C{0.7cm}|C{0.7cm}|C{0.7cm}|C{0.7cm}| } 
 \hline

   TAS &Accuracy &Edit & F1@10 & F1@25 &F1@50\\ \hline
%gemini-2.5-pro (defining Listening) & 55.0& 52.2 &42.4  &48.3    & 64.3    \\ \hline
Gemini-2.5-flash &57.2&37.4&39.0&34.5&25.3  \\ \hline
Gemini-2.5-pro &69.8&51.8&58.2&55.2&45.2  \\ \hline
VLM-based &  67.0& 45.7 &48.3 &43.9 & 31.4 \\ \hline
 
\end{tabular}
}
\end{center}
\vspace{-10mm}
\end{table}

\textbf{Stage \#3: Engagement Classification}.%\cite{brown2020language}, LLaMA \cite{dubey2024llama}, Gemma \cite{team2024gemma}

To quantify the performance of the context-aware engagement classifier, we first evaluate LLMs using the manual dense student action annotation, as shown in \cref{table:comparison_llms_manual_context}.  We can notice that both Llama-3-8B  \cite{dubey2024llama} and Gemma-2-9B \cite{team2024gemma} achieve similar performance in terms of F1 score. The moderate performance of the LLMs on the disengaged class can be attributed to the inherent ambiguity of student engagement. In our dataset, some students were doodling on tablets rather than taking notes; experts labeled these cases as disengaged, but our action labels do not distinguish fine-grained behaviors such as writing versus drawing. Additionally, students who were daydreaming were sometimes annotated as listening because their posture suggested attention despite their lack of focus. Finally, student seated at the classroom periphery may appear to look forward without actually paying attention to the instructor. Since the context-aware engagement classifier (Gemma) achieves the highest F1-score of 92\% using manually annotated actions, we subsequently employ Gemma with automated action parsing. As shown in \cref{table:comparison_llms_vlm_context}, it achieves an F1-score of 86\% with Gemini-based parsing and 88\% with VLM-based parsing. This performance gap is attributable to the difficulty of the student action recognition task, as even the advanced Gemini-2.5-Pro model struggles to accurately parse the 2-minute-long student videos.
%The context-aware engagement classifier achieves 92\% F1 score when using ground truth action annotations. When employing automated action parsing, it attains 86\% with Gemini-based parsing and 88\% with the VLM-based parsing as shown in \cref{table:comparison_llms_vlm_context}. 

\begin{table}
\caption{Comparison of LLMs for context-aware engagement classification based on manual action annotation of student tubelets.}
\vspace{-15pt}
\label{table:comparison_llms_manual_context}
\begin{center}
 {\footnotesize
\begin{tabular}{|C{2cm}|C{1.5cm}|C{0.8cm}|C{1cm}|C{1cm}|}  %|c|c|c|c|c|c|
 \hline
  \;\;\;\;\;\; LLM\;\;\;\;\;\; &&Recall &Precision & F1-score \\ \hline
 
    \multirow{3}{6em}{\mbox{Llama-3-8B \cite{dubey2024llama}}}
   &disengaged & 0.76  &    0.56   &   0.65    \\ 
   &engaged & 0.92  &    0.97   &   0.94    \\\cline{2-5} 
   & weighted avg& \underline{0.90} &\textbf{0.92}     & \underline{0.91}    \\ \hline

   \multirow{3}{6em}{\mbox{Gemma-2-9B \cite{team2024gemma}}}
   &disengaged & 0.69  &    0.67   &   0.68    \\ 
   &engaged & 0.95  &    0.96   &   0.95    \\\cline{2-5} 
   & weighted avg& \textbf{0.92} &\textbf{0.92}     & \textbf{0.92}    \\ \hline

    \multirow{3}{6em}{\mbox{GPT-3.5-turbo \cite{brown2020language} }}
   &disengaged & 0.71  &    0.52   &   0.60    \\ 
   &engaged & 0.91  &    0.96   &   0.93    \\\cline{2-5} 
   & weighted avg& 0.89 &0.90     & 0.89    \\ \hline

\end{tabular}
}
\end{center}
\vspace{-4mm} % Adjust this value as needed
\end{table}

\begin{table}
\caption{Comparison of context-aware engagement classifier based on VLM action parsing and Gemini-2.5-pro-based temporal segmentation of student tubelets using Gemma-2-9B.}
\vspace{-15pt}
\label{table:comparison_llms_vlm_context}
\begin{center}
 {\footnotesize
\begin{tabular}{|C{2cm}|C{1.5cm}|C{0.8cm}|C{1cm}|C{1cm}|}%{|c|c|c|c|c|c|} 
 \hline
 
 Action parsing  &&Recall &Precision & F1-score \\ \hline %of student tubelet
 \multirow{3}{6em}{Gemini-2.5-Pro-based}
&disengaged &  0.58  &  0.43     &   0.49     \\ 
   &engaged & 0.89   &   0.94     &      0.92  \\\cline{2-5} 
   & weighted avg& 0.85 &   \textbf{0.88} &    \underline{0.86}     \\ \hline

\multirow{3}{6em}{\mbox{Our VLM-based}}
&disengaged &  0.53  &  0.52     &   0.52     \\ 
   &engaged & 0.93   &   0.93     &      0.93  \\\cline{2-5} 
   & weighted avg& \textbf{0.88} &   \textbf{0.88} & \textbf{0.88}     \\ \hline

\end{tabular}
}
\end{center}
\vspace{-5mm}
\end{table}
%======================================#

%\subsection{ \textcolor{red}{Comparisons with State-of-the-art}}
\subsection{Comparison with Existing Works}
Direct comparison with prior work is difficult due to differences in datasets, action classes, and annotation protocols. Therefore, we implemented the approach in \cite{abdelkawy2024measuring} by leveraging the predicted actions from the fine-tuned VLM model and applying its reported best-performing classifier. For each student tublet, we obtain the sequence of predicted actions and construct the histogram of actions that represents the frequency of actions within a 2-minute interval. This histogram is then used as an input for the Random Forest classifier for engagement measurement. Since the method in  \cite{abdelkawy2024measuring} requires training, we evaluate it using Leave-One-Out Cross-Validation (LOOCV) on the entire dataset, which contains 3 30-minute lectures. As shown in \cref{tab:existingWorkComparison}, the proposed framework surpasses the compared method by 3\% in F1-score. It also achieves a higher F1-score for the disengaged class (52\% vs. 47\%). This improvement highlights the effectiveness of our approach, which uses a student’s action sequence augmented with the actions of their peers.
%Moreover, the proposed framework clearly outperforms the latter in terms of F1-score for the disengaged class (52\% vs. 47\%). This substantial gap demonstrates the effectiveness of the proposed method that uses a student sequence of actions augmented by his peer sequence of actions. 

\begin{table}
\caption{Comparison with existing work based on the proposed VLM-based action parsing of student tubelets using Gamma-2.9B.}
\vspace{-15pt}
\label{tab:existingWorkComparison}
\begin{center}
{\footnotesize
\begin{tabular}{|c|c|c|c|c|c|} 
 \hline
  \;\;\;\;\;\; Method\;\;\;\;\;\; &&Recall &Precision & F1-score \\ \hline
   \multirow{3}{6em}{\cite{abdelkawy2024measuring} }
   &disengaged & 0.70  &    0.36   &   0.47    \\ 
   &engaged & 0.84  &    0.97   &   0.90    \\\cline{2-5} 
   & weighted avg& 0.83 &0.91     & \underline{0.85}    \\ \hline

\multirow{3}{6em}{Ours}
&disengaged & 0.53  &    0.52   &   \textbf{0.52}    \\ 
   &engaged & 0.93  &    0.93   &   \textbf{0.93}    \\\cline{2-5} 
   & weighted avg&0.88& 0.88  &     \textbf{0.88}     \\ \hline

\end{tabular}
}
\end{center}
\vspace{-8mm} % Adjust this value as needed
\end{table}
\subsection{Cross-Course Evaluation}
Our student engagement dataset, consisting of 3 30-minute lectures, was collected from a single course.
This raises a question: Can the proposed framework generalize to different classes with new students, instructors, and course content? To address this, we collected an additional 30-minute lecture from a separate class involving eight different students. This yielded 113 two-minute segments, consisting of 73 engaged and 40 disengaged samples.  \cref{table:cross_class_eval} presents a comparison of the context-aware engagement classifier based on ground truth manual annotation, Gemini-2.5-Pro-based temporal action segmentation (TAS), and our VLM-based TAS on a 30-minute lecture from a different class. The results show that the proposed framework with VLM-based TAS surpasses the Gemini-2.5-Pro-based variant by a substantial margin of 9\% in weighted F1-score. This improvement highlights the generalization capability of the proposed VLM–LLM framework, demonstrating its ability to accurately parse and classify student action sequences despite variations in student cohorts and student–teacher dynamics.
%We can notice that the proposed framework with our VLM-based TAS outperforms its counterpart one with  Gemini-2.5 Pro-based by a large margin of 9\% in weighted F1-score. This demonstrates the generalization capability of the proposed VLM-LLM framework, in which it can accurately parse and classify the student sequence of predicted actions regardless of variations in student cohort or student-teacher dynamics.

\begin{table}
\caption{Evaluation of a context-aware engagement classifier on a 30-minute lecture from different class using Gemma-2-9B.}
\vspace{-15pt}
\label{table:cross_class_eval}
\begin{center}
{\footnotesize
\begin{tabular}{|C{2cm}|C{1.5cm}|C{1cm}|C{1cm}|C{1cm}|}%{|c|c|c|c|c|} 
 \hline
   Student tubelet annotation &&Recall &Precision & F1-score \\ \hline
   \multirow{3}{6em}{Manual annotation}
   &disengaged & 0.85  &    0.81   &   0.83    \\ 
   &engaged & 0.89  &    0.92   &   0.90    \\\cline{2-5} 
   & weighted avg&\textbf{0.88} &\textbf{0.88}     & \textbf{0.88}    \\ \hline \hline

\multirow{3}{6em}{Gemini-2.5-Pro-based}
&disengaged & 0.90  &    0.56   &   0.69    \\ 
   &engaged & 0.62  &    0.92   &   0.74    \\\cline{2-5} 
   & weighted avg&0.72& 0.79  &     0.72     \\ \hline
   
\multirow{3}{6em}{Our VLM-based}
&disengaged & 0.65  &    0.84   &   0.73    \\ 
   &engaged & 0.93  &    0.83   &   0.88    \\\cline{2-5} 
   & weighted avg&\underline{0.83}& \underline{0.83}  &\underline{0.83}     \\ \hline
   
\end{tabular}
}
\end{center}
\vspace{-8mm} % Adjust this value as needed
\end{table}

\subsection{ Ablation Studies}
To ablate our model, we conducted two experiments: a) removing the classroom context to assess its contribution, and b) comparing our sequence-based method with a histogram-based counterpart. In this section, all engagement classification experiments were performed using Gemma-2.9B, based on the action sequences obtained from our VLM-based action parsing for each 2-minute student video.
%In this section, we conducted experiments to ablate the performance of engagement classifier without classroom context and then evaluate the proposed sequence-based method and its histogram-based counterpart. In these experiments, we use the manual dense temporal action labels for each 2-minute-long student video and Llama-3-8B for engagement classification.

\textbf{a) Engagement classification without classroom context:}
\cref{table:comparison_w_o_context}  shows a comparison between the proposed peer-aware approach and its peer-free counterpart, respectively. It is noticeable that the proposed peer-aware approach outperforms its peer-free counterpart, achieving a 1\% higher F1-score for the disengaged class. Moreover, a context-free approach performs badly ($<50\%$precision) in identifying disengaged students, which is critical. The prompt of the context-free engagement classification is provided in the supplementary material \ref{sec:used_prompts}.%(Sec. 1).
%The prompt for context-free engagement classification is shown in \cref{sec:used_prompts}.%\cref{fig:context_free_prompt}.
\begin{table}
\caption{Comparison of engagement classifier with/out classroom context based on the introduced VLM-based action parsing of student tubelets using Gemma-2-9B.}
\vspace{-15pt}
\label{table:comparison_w_o_context}
\begin{center}
{\footnotesize
\begin{tabular}{|c|c|c|c|c|c|} 
 \hline
  \;\;\;\;\;\; w/o\;\;\;\;\;\; &&Recall &Precision & F1-score \\ \hline

 \multirow{3}{6em}{Context-free }
   &disengaged & 0.62  &    0.44   &   0.51    \\ 
   &engaged & 0.89  &    0.94   &   0.92    \\\cline{2-5} 
   & weighted avg& \underline{0.86}&0.88     &\underline{ 0.87}    \\ \hline
   
     \multirow{3}{6em}{context-aware }
   &disengaged & 0.53  &    0.52   &   0.52    \\ 
   &engaged & 0.93  &    0.93   &   0.93    \\\cline{2-5} 
   & weighted avg& \textbf{0.88}&\textbf{0.88}     &\textbf{ 0.88}    \\ \hline

\end{tabular}
}
\vspace{-4mm} % Adjust this value as needed
\end{center}

\end{table}
 
\textbf{b) Input representation:}
%Here, we compare two types of student action representation: a sequence and a histogram. 
The sequence-based and histogram-based input representations of a student's time-stamped actions are compared in \cref{table:comparison_input_rep}. We note that the proposed sequence-based method achieved an 88\% F1-score compared to 86\% for its histogram-based counterpart. Specifically, the sequence-based approach clearly outperforms the latter in terms of F1-score for the disengaged class (52\% vs. 39\%). This substantial difference shows the student engagement is better captured by the temporal action sequence rather than by aggregated action durations represented as a histogram. Furthermore, temporal action sequences preserve the order and timing of events, enabling relational analysis of a student’s behavior in the context of peers’ concurrent actions. In contrast, action histograms lack temporal structure and therefore fail to capture such contextual dynamics.

%This substantial difference supports our hypothesis that what really reflects student engagement is the temporal sequence of actions rather than an overall duration of actions represented by a histogram. Moreover, temporal action sequence, which enables a relational analysis of engagement by maintaining the order and timing of events, allow our framework to interpret a student's behavior relative to the concurrent actions of the peer 
 %We can observe that sequence-based and histogram-based approach achieve simialr F1 score, especially on disengaged class. However, the the recall of disenaged for sequence-based is 72\% compared to 49\% for histogram-based approach. This confirms our claim that  it is not the total time spent in each action (e.g., via a histogram) but the temporal sequence and context of actions that more accurately reflects engagement. 

\begin{table}
\caption{Comparison of input representation for engagement classification based on the introduced VLM action parsing of student tubelets Using Gemma-2-9B.}
\vspace{-15pt}
\label{table:comparison_input_rep}
\begin{center}
{\footnotesize
\begin{tabular}{|c|c|c|c|c|} 
 \hline
  \;\; Representation\;\; &&Recall &Precision & F1-score \\ \hline
\multirow{2}{3em}{Histogram}
&disengaged & 0.35  &    0.44   &   0.39    \\ 
   &engaged & 0.94  &    0.91   &   0.93    \\\cline{2-5} 
   & weighted avg&0.87& 0.86  &     0.86     \\ \hline

\multirow{2}{3em}{Sequence }
   &disengaged & 0.53  &    0.52   &   0.52    \\ 
   &engaged & 0.93  &    0.93   &   0.93    \\\cline{2-5} 
   & weighted avg& \textbf{0.88} &\textbf{0.88}     & \textbf{0.88}    \\ \hline
   
\end{tabular}
}
\end{center}
\vspace{-8mm} % Adjust this value as needed
\end{table}

%\subsubsection{Identification of effective prompting techniques}
\vspace{-8pt}
\section{Conclusion}
\vspace{-1mm}
We presented a novel context-aware framework to measure student behavioral engagement in a classroom setting. The proposed framework employs a Vision-Language Model (VLM)-based action parser to get a sequence of student actions during an arbitrary time interval and then feeds this student sequence, along with the classroom context (i.e., peers' activities), to a Large Language Model (LLM) for engagement measurement. The experimental results demonstrated that context awareness is crucial to take into account to detect disengaged students because, depending on the context, the same action may indicate engagement or disengagement. On the constructed dataset, we evaluated the proposed framework and obtained engagement classification F1 scores of 92\%, 86\%, and 88\% using manual, Gemini-based, and modified VLM-based temporal action segmentation, respectively. Furthermore, the proposed framework overcomes the challenge of the scarcity of annotated data required to model the diversity of student behaviors.
\vspace{-2mm}
\section{Ethical Statement}
\vspace{-1mm}
This study was approved by IRB \# 19.0513 and informed consent was obtained from all participants.

%\subsection{Ethical Statement}
%%%%%%%%% REFERENCES
{\small
\bibliographystyle{ieee_fullname}
\bibliography{egbib}
}

\clearpage
\newpage

\twocolumn[
  \begin{center}
    \textbf{Context Matters: Peer-Aware Student Behavioral Engagement Measurement via VLM Action Parsing and LLM Sequence Classification} \\
    \vspace{1em}
    Supplementary Material
  \end{center}
]

 % \appendix
 % \section{Appendix}
 \section{LLM Prompts}
 \label{sec:used_prompts}
\begin{figure}[h!]
    \centering
    \includegraphics[width=0.9\linewidth]{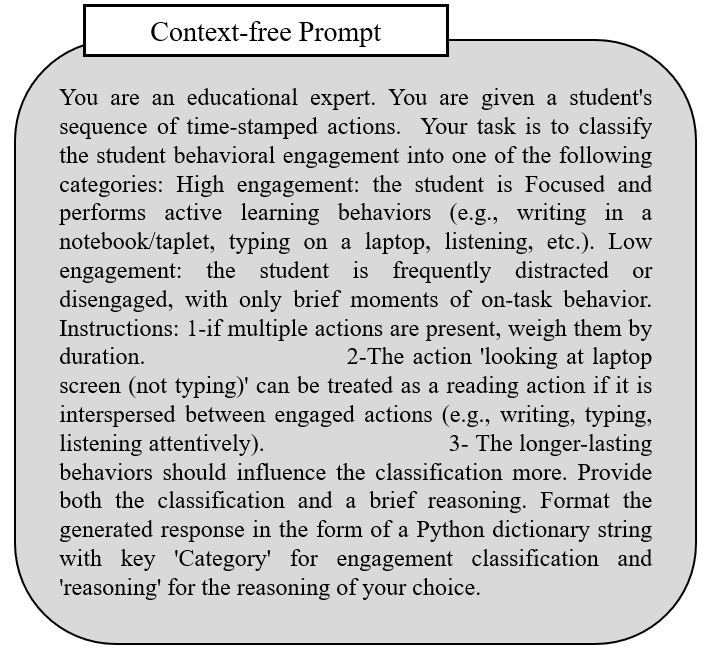}
    \refstepcounter{figure} % Increment figure counter
    \caption{The prompt for context-free engagement classification.}
    \label{fig:context_free_prompt}
\end{figure}

\begin{figure*}[h!]
    \centering
    \includegraphics[width=0.9\linewidth]{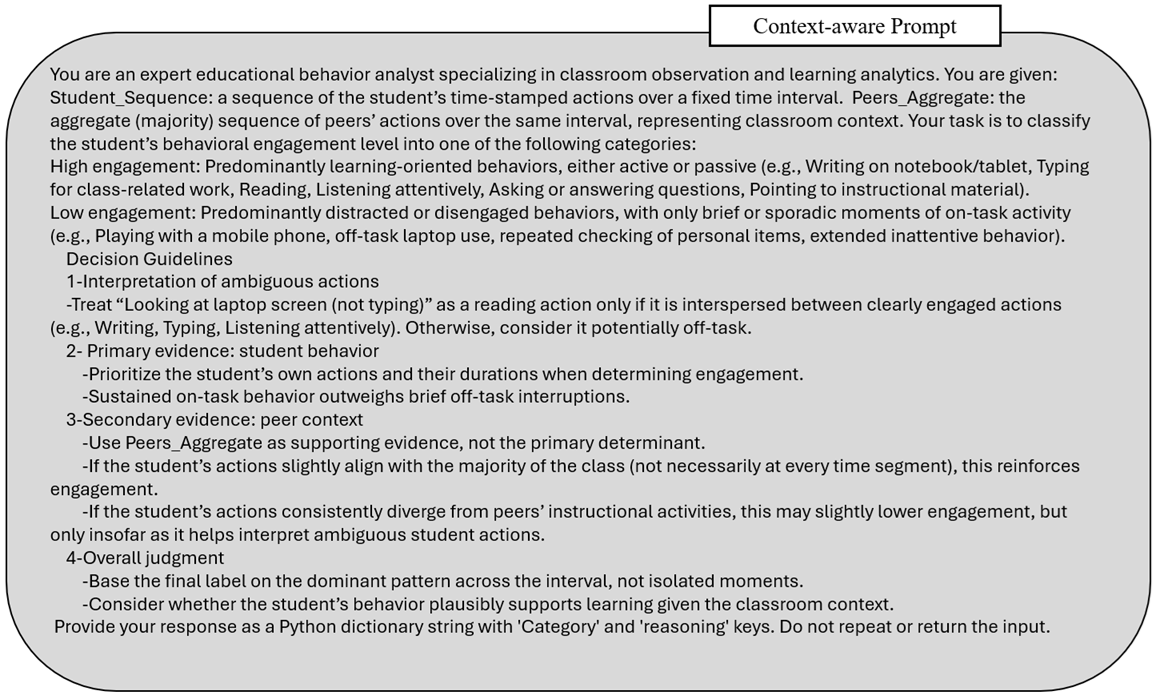}
    \refstepcounter{figure} % Increment figure counter
    \caption{An example of the task description $x_{desc}$}
    \label{fig:context_based_prompt}
\end{figure*}

 In this section, we present the LLM prompts used for:
 \begin{itemize}
    \item Context-free engagement classification (see \cref{fig:context_free_prompt}).
     \item Context-aware engagement classification (see \cref{fig:context_based_prompt}). %with a student's action sequence and its peers' majority sequence as input and output student engagement level and reasoning for such engagement level
 \end{itemize}
The context-free engagement classification uses only a student’s action sequence as input and outputs the student’s engagement level along with the reasoning behind that assessment. In contrast, context-aware engagement classification takes both a student’s action sequence and the majority action sequence of their peers as input and outputs the student’s engagement level together with the reasoning for that assessment. 
%  \subsection{context-aware prompt}
% \label{sec:context_based_prompt}

 % \subsection{Context-free engagement classification prompt}
% \label{sec:context_free_classifier_prompt}
%\texttt{"You are an educational behavior analyst. A student’s classroom behavior is represented as a sequence of time-stamped actions. Your task is to classify the student’s overall behavioral engagement level during the observed interval. Only use the following categories: High engagement: Focused and active learning behaviors (e.g., taking notes, listening attentively, asking/answering questions, typing academically). Low engagement: Frequently distracted or disengaged with only brief moments of on-task behavior. If multiple actions are present, weigh them by duration. The longer-lasting behaviors should influence the classification more. Provide both the classification and a brief reasoning. The generated response should be in the form of a Python dictionary string with key 'Category' for engagement category and 'reasoning' for the reasoning of your choice. Please do not repeat or return the content back again. The student's action(s) with start and end times (in MM:SS format) is/are listed as follows: \{student-action-seq\}"}

\section{MLLM Prompt }
\label{sec:TAC_prompt}
\begin{figure}[h!]
    \centering
    \includegraphics[width=0.9\linewidth]{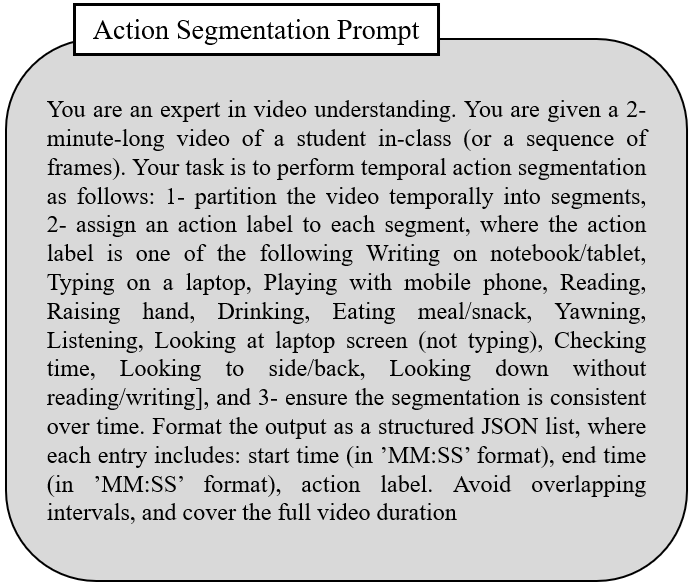}
    \refstepcounter{figure} % Increment figure counter
    \caption{The prompt for temporal action segmentation.} 
    \label{fig:action_segm_prompt}
\end{figure}

To assess the capability of advanced multimodal large language models, e.g., Gemini-2.5 Pro, in parsing a student's actions within a 2-minute video, we designed a prompt that instructs the MLLM to perform temporal action segmentation as shown in \cref{fig:action_segm_prompt}.
 %To assess the advanced multimodal large language model capability on parsing a student's actions within a 2-minute video, we curated a prompt for MLLMs to perform temporal action segmentation.

\end{document}